\documentclass[journal]{IEEEtran}

\usepackage[pdftex]{graphicx}
\DeclareGraphicsExtensions{.pdf,.jpeg,.png}

\usepackage{algorithmic}
\usepackage{booktabs}
\usepackage{multirow} 
\usepackage{url}
\usepackage{hyperref}
\hypersetup{
    colorlinks = true,
    citecolor = blue,
}
\usepackage{array} 
\usepackage{multirow}
\usepackage{longtable}
\usepackage{makecell}
\usepackage{amssymb}
\usepackage{amsmath}

\begin{document}

\title{Explainable Earth Surface Forecasting\\under Extreme Events}

\newcommand{\dataset}{DeepExtremeCubes}
\newcommand{\giturl}{https://github.com/DeepExtremes/txyXAI}

\author{Oscar J. Pellicer-Valero, %
        Miguel-Ángel Fernández-Torres, %
        Chaonan Ji, %
        Miguel D. Mahecha, %
        and~Gustau~Camps-Valls%
\thanks{Oscar J. Pellicer-Valero, Miguel-Ángel Fernández-Torres, and Gustau Camps-Valls are with the Image Processing Laboratory (IPL), Universitat de València, València, 46980, Spain.}%
\thanks{Chaonan Ji and Miguel D. Mahecha are with the Remote Sensing Centre for Earth System Research (RSC4Earth), Leipzig University, Leipzig, 04103, Germany, and the Institute for Earth System Science and Remote Sensing, Leipzig University, Leipzig, 04103, Germany.}%
\thanks{Miguel D. Mahecha is also with the Image and Signal Processing Group, Leipzig University, Leipzig, 04109, Germany, and the Helmholtz Centre for Environmental Research -- UFZ, Leipzig, 04318, Germany.}%
}

\markboth{}%
{Pellicer-Valero \MakeLowercase{\emph{et al.}}: xAI for Earth Surface Forecasting under Extremes}

\maketitle

\begin{abstract}

With climate change-related extreme events on the rise, high dimensional Earth observation data presents a unique opportunity for forecasting and understanding impacts on ecosystems. This is, however, impeded by the complexity of processing, visualizing, modeling, and explaining this data. To showcase how this challenge can be met, here we train a convolutional long short-term memory-based architecture on the novel \dataset~dataset. \dataset~includes around 40,000 long-term Sentinel-2 minicubes (January 2016-October 2022) worldwide, along with labeled extreme events, meteorological data, vegetation land cover, and topography map, sampled from locations affected by extreme climate events and surrounding areas. When predicting future reflectances and vegetation impacts through kernel normalized difference vegetation index, the model achieved an R$^2$ score of 0.9055 in the test set. Explainable artificial intelligence was used to analyze the model's predictions during the October 2020 Central South America compound heatwave and drought event. We chose the same area exactly one year before the event as counterfactual, finding that the average temperature and surface pressure are generally the best predictors under normal conditions. In contrast, minimum anomalies of evaporation and surface latent heat flux take the lead during the event. A change of regime is also observed in the attributions before the event, which might help assess how long the event was brewing before happening. The code to replicate all experiments and figures in this paper is publicly available at \url{\giturl}.
\end{abstract}

\begin{IEEEkeywords}
Earth surface forecasting, deep learning (DL), explainable artificial intelligence (xAI), kernel normalized difference vegetation index (kNDVI), vegetation impacts, convolutional long short-term memory (ConvLSTM)
\end{IEEEkeywords}

\section{Introduction}

\begin{figure}[!t]
    \centering
    \includegraphics[width=\columnwidth]{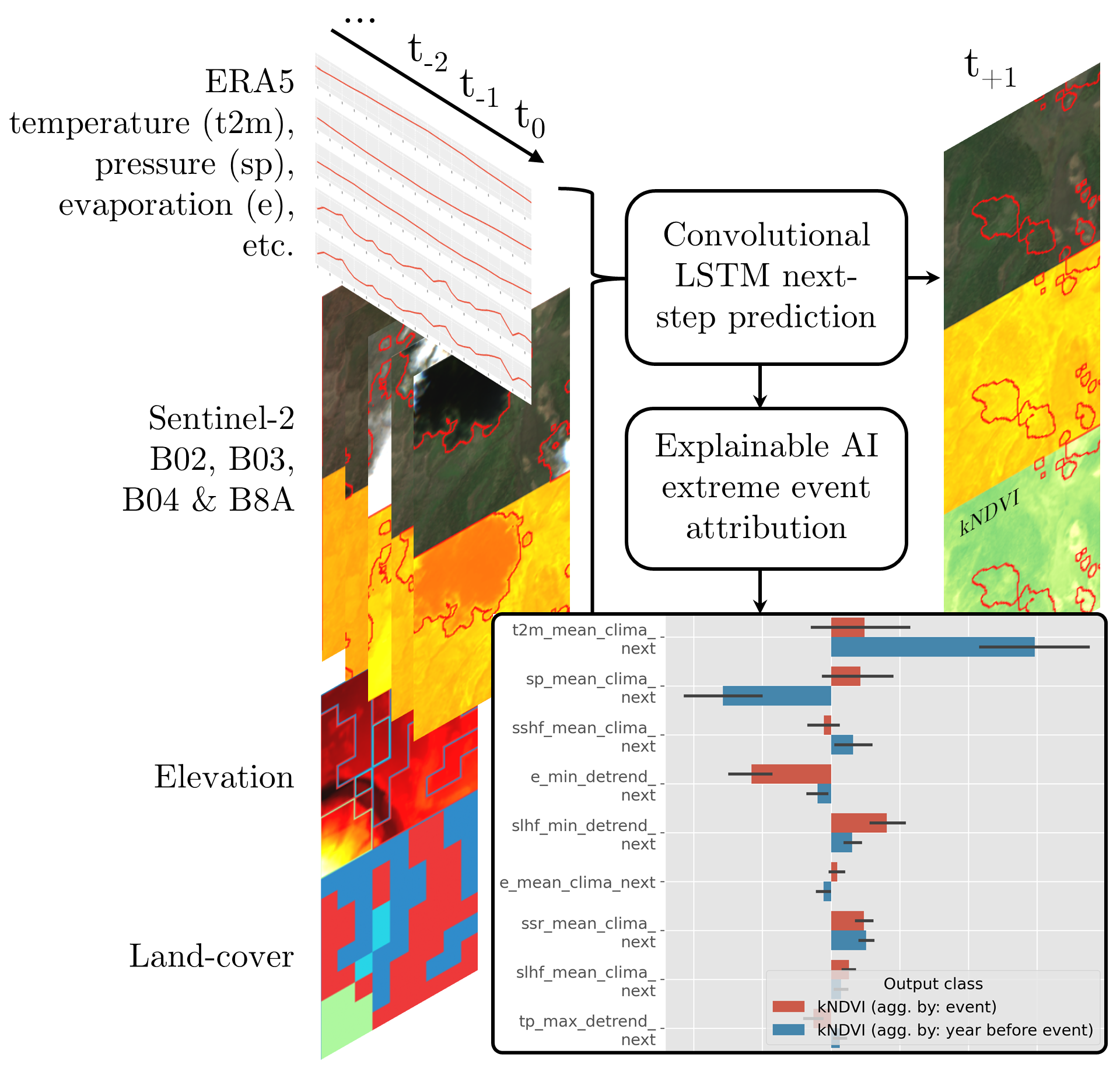}
    \caption{A convolutional LSTM model was trained to forecast future Sentinel-2 reflectances and vegetation impacts given previous timesteps (augmented with ERA5 meteorology, elevation, and land cover). Explainable AI was then used to gather insights into the effects of extreme events on vegetation by comparing the model's attributions during event and comparable non-event situations. kNDVI: kernel normalized difference vegetation index.}
    \label{fig:pipeline}
\end{figure}

\IEEEPARstart{C}{limate} change is amplifying the frequency and intensity of extreme weather events, which in turn disrupt ecosystem functioning, diminishing carbon sequestration \cite{Reichstein2013}, water retention \cite{Terrado2014}, and biodiversity \cite{Mahecha2024}. All this contributes to harvest failures, directly affecting human well-being. Multi-hazard events often manifest as compound events \cite{Zscheischler2020}, posing a much greater impact on society and the environment than individual events. For instance, a Compound Heatwave and Drought (CHD) event can result in global food production issues \cite{Gaupp2019}. The ability to predict and understand the genesis and drivers of compound events could enable the effective implementation of mitigating measures by governmental bodies \cite{UNEP2020}. Applying explainable Artificial Intelligence (xAI) methods to Deep Learning (DL) models trained on Remote Sensing (RS) data offers a theoretical possibility to predict and analyze event impacts. 



Early examples of forecasting models for RS data (e.g., predicting satellite reflectance some timesteps ahead) and vegetation impact monitoring (e.g., predicting some vegetation impact index) include \cite{Hong2017}, which used a convolutional Long Short-Term Memory (convLSTM) network \cite{Shi2015} on weather satellite COMS-1 data; or \cite{Xu2019}, in which Generative Adversarial Networks (GANs) \cite{Goodfellow2014} over LSTM networks were applied on FY-2E satellite cloud maps. These forecasting models are generally trained in a self-supervised manner, using future data as prediction targets, eliminating the need for manual labeling and allowing for the collection of arbitrarily large RS datasets. 

Requena-Mesa et al. \cite{RequenaMesa2021} made a substantial contribution to the field by providing the EarthNet21 dataset along with some baseline models. These data were specifically tailored to the task of spatio-temporal high-resolution meteorology-guided Earth surface forecasting. It contained around 32,000 Sentinel-2 spatio-temporal arrays (also known as ``minicubes''), along with a static Digital Elevation Model (DEM). In the same vein, a convLSTM was trained to predict future Normalized Difference Vegetation Index (NDVI) values, commonly used as a proxy for vegetation health states, conditioned on past reflectances and future ERA5 atmospheric reanalysis data \cite{Hersbach2020}. Various follow-up works emerged since \cite{Diaconu2022,Kladny2024}. For example, in \cite{Gao2022}, Gao et al. developed a transformer-based architecture that improved the EarthNet21 metrics, Robin et al. \cite{Robin2022} focused on the African continent and, very recently, Benson et al. \cite{Benson2024} proposed GreenEarthNet, a new dataset with improved cloud mask, better model baselines (using a transformer), and evaluation. Overall, the main limitations of EarthNet21-like approaches are the short context window (50 days, with 5-day period, for a total of ten samples), the locality of the data (both in the sense that only Europe was considered and that only past data from the same location is used as context), and the non-causality of the meteorological data used for the conditioned prediction (which is assumed to be available up to 100 days in the future at ERA5-like accuracy).

Parallel to these developments, a wave of foundation models \cite{Bommasani2021} have started to appear in the field of RS. In general, these models are characterized by using a wide variety of input modalities, such as Sentinel-2, Sentinel-1, ERA5 variables, DEM, and land-covers, and employing a self-supervised pretraining objective such as masked modeling, wherein pixels/patches, channels/bands, or even complete timesteps are removed from the inputs. The model has to learn to reconstruct them, such as in \cite{Tseng2024}. After training, the decoder part from the pretrained masked autoencoder is dropped, and the encoder part can then be applied (either by linear-probing the encoder's output or by fine-tuning the encoder's weights) to a variety of downstream tasks \cite{Cong2022,Sun2022,Reed2023,Smith2024}. In all the previous examples, the authors improve the performance compared to training a model directly on the downstream task. However, smaller datasets seem to benefit the most from the pretraining. All analyzed methods are unfortunately limited to static satellite images (do not consider the time dimension) and the modalities seen during training (e.g., cannot adapt to other sensors), showing visually degraded reconstructions for the masked areas. 


One of the downsides of these highly-dimensional models is their lack of transparency (i.e., knowing their weights does not help us understand how they work). The field of xAI was therefore born to make them interpretable (i.e., help us understand their decisions) and explainable (i.e., link these interpretations to domain knowledge) \cite{Roscher2020}. One of the most successful sub-fields in xAI is feature attribution, which attempts to assess the impact of an input feature (either positive or negative) on the prediction outcome \cite{Molnar2023}. For instance, an LSTM model is trained in \cite{MateoSanchis2023} to predict crop yield, being the attribution method Shapley Additive Explanations (SHAP) \cite{Lundberg2017} used to conclude that high temperatures during the growth season have a negative impact on crop yield or to discover the existence critical periods of the crop growth cycle for the corn, soybean, and wheat models. State-of-the-art (SOTA) convolutional Neural Networks (CNNs) for the BigEarthNet \cite{Sumbul2019} and SEN12MS \cite{Schmitt2019} datasets were explained in \cite{Kakogeorgiou2021} with ten different xAI methods, selecting Grad-CAM \cite{Selvaraju2016}, Occlusion, and Lime \cite{Ribeiro2016} as the best for their purposes. Almost no authors, however, have applied xAI to spatio-temporal models for RS data. A single example was found: Huang et al. \cite{Huang2023} trained a convLSTM model for soil moisture prediction in China from ERA5 data and then used Permutation Importance \cite{Altmann2010} for global (i.e., model-wide) interpretations and Smooth Gradient \cite{Smilkov2017} for local ones (i.e. sample-wise), the latter based both on temporal and spatial aggregations (medians). For a recent and exhaustive review on xAI for RS, refer to \cite{hohl2024xai}. As can be appreciated, the literature has a considerable gap regarding xAI application and visualization on highly dimensional spatio-temporal models. The challenge is furthered by the fact that very compute-intensive xAI methods such as Occlusion or Lime cannot be used in practice for such data.

The literature on using xAI to explain extreme events such as heatwaves and droughts is quite scarce. XGBoost was used in \cite{Mardian2023} to predict the Canadian Drought Monitor severity index given many manually crafted features and indices; then, the average SHAP values were used to obtain global feature attribution. Similarly, Mardian et al. \cite{Mardian2023} used an LSTM over climatic variables, such as El Niño-Southern Oscillation (ENSO) over different regions, to forecast the Standard Precipitation Index (SPI), using average SHAP and Partial Dependence Plots (PDP) for global attribution. xAI has also been applied to understand other events such as rainfall extremes \cite{Rampal2022}, river flooding \cite{Jiang2022}, or wildfires \cite{Kondylatos2022}. Li et al. \cite{Li2024} represent SHAP attributions for wildfires over spatial maps, indicating the most important features of each region and leading to interesting insights. No papers were found on using xAI for compound heatwaves and drought events.


Our main contributions, illustrated in Figure \ref{fig:pipeline}, are as follows: First, we provide a comprehensive toolset for processing and visualizing spatio-temporal data, as well as for training and evaluating DL models, successfully applying it to the novel very long-context \dataset~dataset. In particular, regarding data processing, we introduce an efficient technique for causal climatology computation while, from a modeling standpoint, we ensure consistent prediction of both reflectances and the kernel NDVI vegetation index via a novel loss function. Second, we offer tools for applying xAI to high-dimensional data, with visualizations at both local (single input) and global levels, demonstrated through the analysis of the October 2020 central South America heatwave. To our knowledge, this is the first publication detailing DL training on global long-context, high-resolution remote sensing data, as well as the first to apply xAI to such data and visualize the results. The code for replicating all experiments and figures is publicly available at \url{\giturl}.

\section{Materials and methods}\label{sec:materials-and-methods}

\begin{figure*}[!tb]
    \centering
    \includegraphics[width=\linewidth]{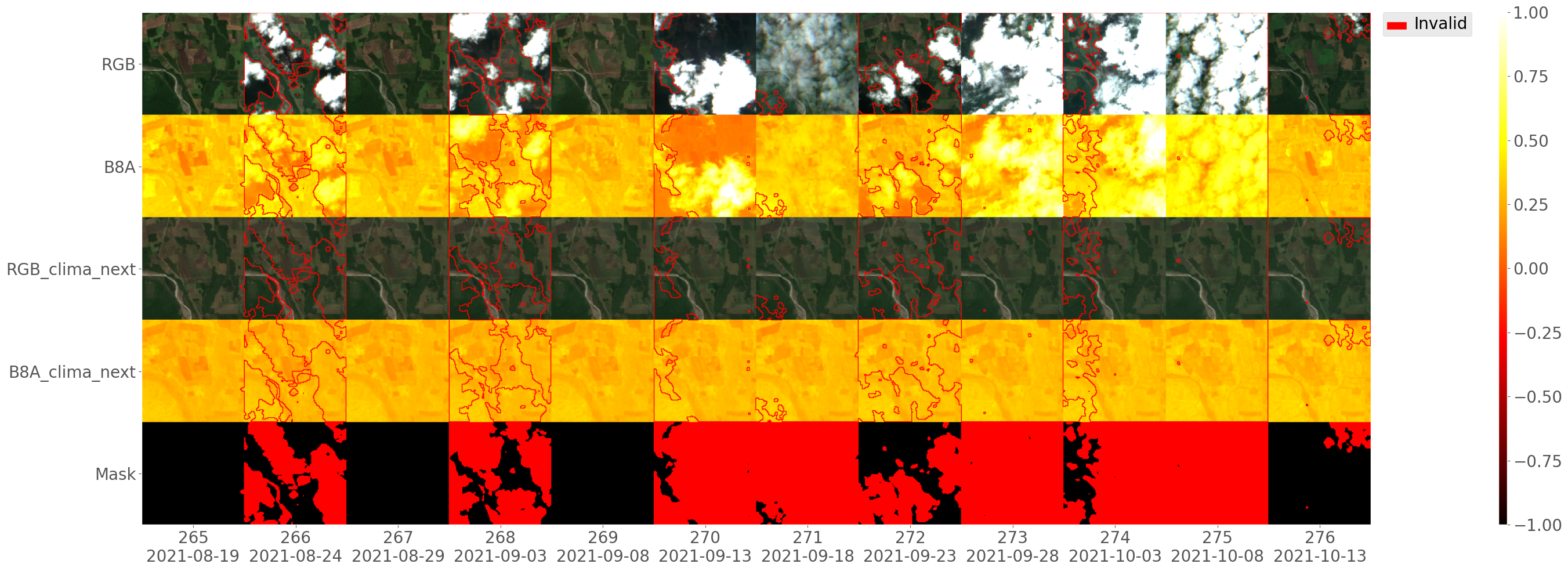}
    \caption{Overview of the spatio-temporal inputs (tensor $x_{st}$) for a single minicube sampled at location -99.47, 24.17 (China). Only a few timesteps are shown from 2021-08-19 to 2021-10-13.}
    \label{fig:txy}
\end{figure*}

\subsection{Dataset and data preprocessing}\label{dataset}

The \dataset~dataset was used for conducting all the experiments. It contains $\sim$40.000 128 pixels $\times$ 128 pixels, globally sampled small data cubes (i.e. minicubes), with a spatial coverage of 2.5 by 2.5 km. Each minicube includes (i) Sentinel-2 L2A images, (ii) ERA5-Land variables \cite{munoz2021era5} and generated extreme event cube covering 2016 to 2022 \cite{weynants_dheed_paper_2024}, and (iii) ancillary land cover and topography maps \cite{CopernicusDEM}. For more information on this dataset, please refer to \cite{ji2024deepextremecubes}. The variables in each minicube were preprocessed into three large tensors according to their dimensionality (see Table~\ref{tab:features} for a summary):




\begin{table*}[!tb]
    \centering
    \renewcommand{\arraystretch}{1.2}
    \caption{Preprocessing of DeepExtremeCubes variables to obtain spatio-temporal ($x_{st}$), spatial ($x_s$), and temporal ($x_t$) input features}

    \begin{tabular}{p{0.16\linewidth}p{0.12\linewidth}p{0.15\linewidth}p{0.45\linewidth}}
        \makecell{Dimensionality} & 
        \makecell{Original variables} & 
        \makecell{Processed Features} & 
        \makecell{Description} \\
        
        \toprule
        
        \multirow{3}{*}{\makecell[l]{Spatio-temporal\\$ x_{st} \in \mathbb{R}^{C_{st} (9) \times T \times W \times H} $ }}
        & \makecell[l]{\emph{B02}, \emph{B03}, \emph{B04}, \\ \emph{B8A}}
        & \makecell[l]{\emph{B02\_clima\_next}, \\ \emph{B03\_clima\_next}, \\ ... (4 features in total) }
        & \makecell[l]{
            \emph{*\_clima}: time-of-year average of \emph{*}, computed causally\\
            \emph{*\_next}: data refers to the next timestep (+ 5 days)
            }\\
        \cline{2-4}
        & \makecell[l]{\emph{cloudmask\_en}} & \emph{cloud\_mask} & \emph{cloud\_mask} has a value of 1 for \emph{cloudmask\_en}'s classes: cloud, cloud\_shadows,  and masked\_other\_reasons \\
        
        \midrule
        
        \multirow{2}{*}{\makecell[l]{Spatial\\$ x_{s} \in \mathbb{R}^{C_s (34) \times W \times H} $ }} 
        & \makecell[l]{\emph{LCCS\_land\_cover}} & 
        \makecell[l]{\emph{LCCS\_cropland\_rainfed}, \\ \emph{LCCS\_mosaic\_cropland}, \\ ... (33 features in total)} & \makecell[l]{One-hot encoded all 34 \emph{LCCS\_land\_cover}'s classes into 33 \\ new variables (class no\_data was ignored)} \\
        \cline{2-4}
        & \emph{cop\_dem} & \emph{cop\_dem} & No special preprocessing \\
        
        \midrule

        \multirow{2}{*}{\makecell[l]{Temporal\\$ x_{t} \in \mathbb{R}^{C_t (24) \times T } $}} 
        & \makecell[l]{\emph{e}, \emph{pev}, \emph{slhf}, \emph{sp}, \\ \emph{sshf}, \emph{ssr}, \emph{t2m}, \emph{tp} }
        & \makecell[l]{\emph{e\_min\_detrend\_next}, \\ \emph{e\_max\_detrend\_next}, \\ \emph{e\_mean\_clima\_next}, \\ \emph{pev\_min\_detrend\_next}, \\... (24 features in total)}
        & \makecell[l]{
            \emph{*\_min}: minimum aggregation of \emph{*} over the previous 5 days \\
            \emph{*\_max}: maximum aggregation of \emph{*} over the previous 5 days \\
            \emph{*\_mean}: mean aggregation of \emph{*} over the previous 5 days \\ 
            \emph{*\_clima}: time-of-year average of \emph{*}, computed causally \\
            \emph{*\_detrend}: detrended \emph{*}, computed by subtracting \emph{*\_clima} from \emph{*}\\
            \emph{*\_next}: data refers to the next timestep (+ 5 days)
            }\\
        
        \bottomrule
        \label{tab:features}
    \end{tabular}
\end{table*}

\subsubsection{Spatio-temporal data}\label{sec:txy} represented by tensor $ x_{st} \in \mathbb{R}^{C_{st} \times T \times W \times H}$ with dimensions $ C_{st} \times T \times W \times H $, being $C_{st}=9$ channels, $T=495$ timesteps (five-day period), and $W=H=$ 128 pixels (at 30m per pixel). It contains a variety of variables (see example minicube in Figure~\ref{fig:txy}):
  \begin{itemize}
      \item Spatio-temporal dynamic satellite reflectance data (\emph{B02}, \emph{B03}, \emph{B04}, and \emph{B8A} Sentinel-2 bands) at 30 m \(\times\) 30 m \(\times\) five days resolution.
      \item Their climatologies for the next timestep (\emph{B02\_clima\_next}, \emph{B03\_clima\_next}, \emph{B04\_clima\_next}, and \emph{B8A\_clima\_next}, to be explained hereafter).
      \item The \emph{cloud\_mask}: with segmentation masks for clouds, cloud shadows, snow, and cloud-free pixels. Instead of doing imputation of the cloud-covered regions, the cloud mask was fed directly as an input so the model could learn to deal with the missing data.
  \end{itemize}

The computation of the climatology requires further explanation. For each spatial location, climatology refers to the average value of a variable for a given day of the year based on previous years of observations. For our purposes, climatology was computed by splitting the year into monthly bins, averaging all observations falling within each bin, and then interpolating between bins to get the final values for a given time of the year. This process is efficient, as it only requires a linear interpolation between two values for a given timestep; causal, as, for a given timestep, it only uses data that has been observed up to that point, which is crucial for deployment in a real-time scenario; and dynamic, since it constantly gets updated as new samples are observed. Also, the climatology was computed for the next timestep (the one the model will attempt to predict), hence helping the model in its task. This is reflected in the name of these variables (\emph{*\_clima\_\bfseries{next}}). 
  
\subsubsection{Spatial data}\label{sec:xy} represented by tensor $x_{s} \in \mathbb{R}^{C_s \times H \times W}$, with $C_s=34$ channels. It contains:
\begin{itemize}
    \item A one-hot encoding of the \href{https://www.fao.org/3/x0596e/x0596e01f.htm}{LCCS vegetation-focused land cover}, with a total of 34 different classes (downscaled from 240 m \(\times\) 240 m resolution to 30 m \(\times\) 30 m using nearest-neighbor interpolation).
    \item The \href{https://spacedata.copernicus.eu/collections/copernicus-digital-elevation-model}{Copernicus DEM} (\emph{cop\_dem}), which has the same 30 m native resolution as the spatio-temporal data.
\end{itemize}

  
\subsubsection{Temporal data}\label{sec:t} represented by tensor $x_{t} \in \mathbb{R}^{C_t \times T}$ with $C_t=24$ channels, containing exclusively ERA5-Land-derived variables: 
  \emph{e} (\href{https://apps.ecmwf.int/codes/grib/param-db?id=182}{evaporation}, m of water equivalent), 
  \emph{pev} (\href{https://apps.ecmwf.int/codes/grib/param-db?id=228251}{potential evaporation}, m),
  \emph{slhf} (\href{https://apps.ecmwf.int/codes/grib/param-db?id=147}{surface latent heat flux}, J/m$^2$),
  \emph{sp} (\href{https://apps.ecmwf.int/codes/grib/param-db?id=134}{surface pressure}, Pa),
  \emph{sshf} (\href{https://apps.ecmwf.int/codes/grib/param-db?id=146}{surface sensible heat flux}, J/m$^2$),
  \emph{ssr} (\href{https://apps.ecmwf.int/codes/grib/param-db?id=176}{surface net short-wave (solar) radiation}, J/m$^2$),
  \emph{t2m} (\href{https://apps.ecmwf.int/codes/grib/param-db?id=130}{temperature at 2m}, $^{\circ}$K),
  \emph{tp} (\href{https://codes.ecmwf.int/grib/param-db/?id=228}{total precipitation}, m). 
These variables, available at a 6-hourly rate, had to be aggregated to match the five-day spatio-temporal data. Several aggregation strategies were applied (* stands for any of the above variables):
\begin{itemize}
\item \emph{*\_min\_detrend\_next}: Minimum anomaly (with respect to climatology) over the following five days.
\item \emph{*\_max\_detrend\_next}: Maximum anomaly (with respect to climatology) over the following five days.
\item \emph{*\_mean\_clima\_next}: The mean climatology value over the following five days.
\end{itemize}

Note that the variables include \emph{\_next} in their name, meaning that their values refer to the next timestep. Therefore, this is a non-causal ERA5 usage, assuming that in a real-time scenario, ERA5-quality atmospheric forecasts would be available up to 5 days in the future. For comparison, similar works  \cite{RequenaMesa2021,Benson2024} follow the much stronger assumption that such forecasts would be available for up to 100 days.

\subsubsection{Data standardization}\label{sec:estandardization}
For the Sentinel-2 bands, data was left unmodified; for the \emph{cop\_dem}, data was divided by the maximum height in Earth (8849 m); for the ERA5-variables, 0.01\% and 99.99\% percentiles were computed on the training set values, and variables ($v$) were standardized ($\hat{v}$) by following
\begin{equation*}
    \hat{v}=\frac{v-{\mathrm{perc}}(v,0.01\%)}{{\mathrm{perc}}(v,99.99\%)-{\mathrm{perc}}(v,0.01\%)}
\end{equation*}
where $perc$ is the percentile operation. Finally, to get rid of any remaining large values that could destabilize training, all data was additionally clipped to the $[-5,5]$ range.

\begin{figure*}[!tb]
   \centering
   \includegraphics[width=\linewidth]{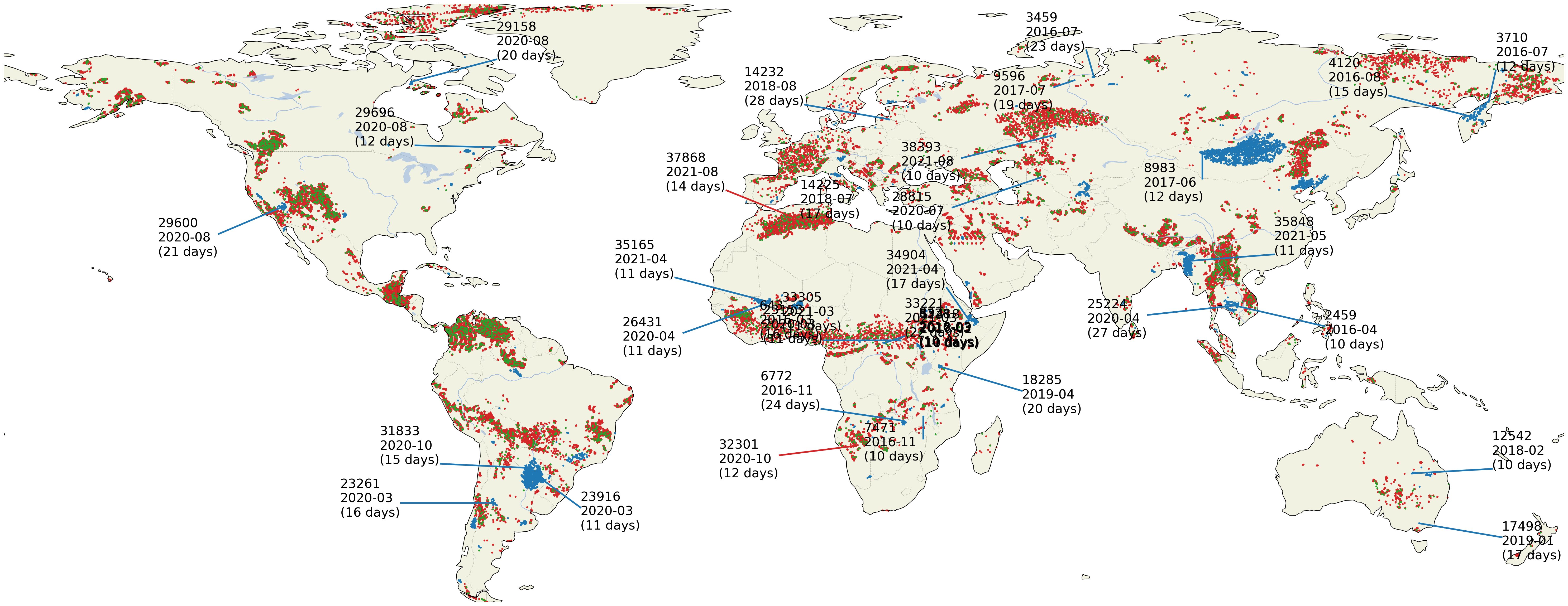}
   \caption{Map of the geographical distribution of the minicubes according to the subset to which they belong (red: train, green: validation, blue: test). Also, events lasting at least ten days have been marked on the map, along with their label ID, occurrence time, and duration.}
   \label{fig:map}
\end{figure*}

\subsection{Earth surface and kNDVI forecasting model}\label{sec:forecasting}

\subsubsection{Task}\label{sec:task}
A convLSTM was trained on this data ($x_{st}, x_s$ and $x_t$) for the task of vegetation impact forecasting through the prediction of kernel NDVI (kNDVI), a common proxy for vegetation health state. In particular, the model was trained to predict the same Sentinel-2 input bands, but one timestep (five days) into the future (outputs: \emph{B02\_next},\emph{ B03\_next}, \emph{B04\_next}, and \emph{B8A\_next}). From these predicted output bands, kNDVI was computed by following Equation \ref{eq:kndvi} (where $\epsilon = 1 \cdot 10^{-5}$ to avoid division by zero). This self-supervised task might seem simple, but in practice, the model must often effectively predict many more days into the future, as many minicubes are almost perpetually covered by clouds (in the training set, more than 50\% of the minicubes have at least a 50\% cloud coverage), making the problem much harder.
\begin{equation}
    \mathrm{kNDVI} = \tanh{ \left(\frac{\mathrm{B8A} - \mathrm{B04}}{\mathrm{B8A} + \mathrm{B04} + \epsilon}\right)^2}
    \label{eq:kndvi}
\end{equation}

Following the idea proposed by \cite{Benson2024}, model inputs \emph{B02\_clima\_next}, \emph{B03\_clima\_next}, \emph{B04\_clima\_next}, and \emph{B8A\_clima\_next} were added to the predicted outputs to obtain the final \emph{B02\_next}, \emph{B03\_next}, \emph{B04\_next}, and \emph{B8A\_next}. Hence, the model is forced to predict the difference with respect to the climatology (i.e., the anomalies) instead of wasting modeling capacity in reconstructing the full reflectances. Besides, the training becomes much more stable as anomalies are typically zero-centered and small in variance.

\subsubsection{Architecture}\label{sec:architecture}
The employed architecture is a standard 3-layer convLSTM with a total of 768k parameters implemented using PyTorch Lightning~\cite{pytorchlighning2019}. Before reaching the network, tensors $x_{st}, x_s$, and $x_t$ are concatenated along the channel dimension by repeating tensor $x_t$ along all spatial locations and tensor $x_s$ along all temporal locations. These tensors are concatenated with the LSTM's hidden state before going through a single 2D convolution (kernel size $3 \times 3$, padding $2 \times 2$, hidden dimension 60). They are then split into the input, input gate, output, and forget tensors to go through the standard LSTM pathways \cite{Hochreiter1997}. This output is fed to the next convLSTM layer. At the output of the third layer, a pixel-wise two-layer perceptron adapts the hidden dimensionality (60) to the output dimensionality (4) with a single 10-unit hidden layer with ReLU activation. 

Two more models were tested as ablations: \emph{LSTM} (495k parameters), where $3 \times 3$ convolutions were replaced by $1 \times 1$ convolutions (so that the model behaved as a pixel-wise standard LSTM network with no spatial context), and the hidden dimension was increased to try to match the number of parameters of the convLSTM (it could not be matched precisely because of memory limitations); and \emph{Conv} (744k parameters), in which the three convLSTM cells were replaced by three consecutive $3 \times 3$ convolutions, with the number of intermediate channels chosen to approximately match the original model in parameter count.

\subsubsection{Loss}\label{sec:loss}
The L1 loss was used for training, which generally preserves high-frequency details better than the more widely used L2 loss. However, this loss was not directly applied; instead, kNDVI was first computed from network outputs \emph{B04\_next} and \emph{B8A\_next}. Then the L1 loss was applied to the different outputs independently (B02\_next, B03\_next, B04\_next, B8A\_next, and kNDVI\_next), and their respective losses were aggregated (by weighted sum) with weights 0.125, 0.125, 0.125, 0.125, and 0.5 respectively. Hence, the network is encouraged to accurately predict \emph{kNDVI\_next} (as the relative weight is much higher), not by directly predicting it, but rather by accurately predicting the actual reflectance values from which it is calculated. We argue that this is a better training objective than directly predicting derived maps such as kNDVI since the model needs not waste capacity in learning the mapping between reflectances and kNDVI; yet, by propagating the loss through the kNDVI, we directly encourage good kNDVI predictions, which is of high interest for the analysis of vegetation impacts.

Unlike in previous impact assessment studies using NDVI, in this work, we chose kNDVI as our main study output (as compared with NDVI) for two main reasons: Firstly, it is a better proxy to canopy structure, leaf pigment content (and, subsequently, plant photosynthetic potential), and correlates better with Gross Primary Production and Solar-Induced Chlorophyll Fluorescence than other indices (such as NDVI and Near-Infrared Reflectance of Vegetation -NIRv-) at many spatial and temporal scales \cite{CampsValls2021}. Secondly, it allows for the direct optimization of the model with respect to this index. In comparison, propagating the loss directly through the NDVI formula leads to unstable training due to the small denominators leading to arbitrarily large NDVI values. In contrast, the $\tanh$ in kNDVI likely provides a smoothing effect stabilizing training as it bounds predictions.

Finally, the loss was masked by the cloud mask, such that any pixels containing clouds, cloud shadows, unavailable, or otherwise erroneous did not contribute to the training. This led to a model that never predicts clouds, as the loss never rewards it for doing so.

\subsubsection{Train, validation and test subsets}\label{sec:subsets}
For validation purposes, the minicubes were split into two sets: \emph{train+val} and \emph{test}, such that the locations for any given cube in \emph{train+val} were at least 50 km away from all minicube locations in \emph{test}. The dataset was then split into the following subsets:

\begin{itemize}
\item
  Training set: random 80\% of the minicubes from \emph{train+val}, from January 2018 until December 2021.
\item
  Validation set: remaining 20\% of the minicubes from \emph{train+val}, from January 2022 until October 2022. Therefore, these minicubes are temporally uncorrelated from those in the training set but geographically close.
\item
  Test set: all cubes from \emph{test}, from January 2022 until October 2022. The cubes in this set are both spatially ($>$50km away) and temporally uncorrelated with the training set.
\end{itemize}

Figure \ref{fig:map} shows a map of the geographical distribution of the minicubes according to their subset. Note that the first two years of data were not included in any of the subsets: 2016 was removed because only one of the two Sentinel-2 satellites was online during this first year of the mission, and 2017 was removed because at least a year of observations was needed to compute the climatology (as explained in Section~\ref{dataset}).

\subsubsection{Training}\label{sec:training}
The model was trained for four epochs with a batch size of one (the biggest one that would fit in GPU memory) using batch gradient accumulation to keep the effective batch size fixed at eight. The AdamW optimizer with a fixed learning rate of 0.001 was used for optimization. All the decisions regarding the specific values for the different hyperparameters (such as learning rate, number of layers, training schedule, etc.) were based on extensive experimentation on the validation set. Every epoch took around 16 hours on a high-end A100-SXM4-80GB GPU.

\subsubsection{Evaluation}\label{sec:evaluation}
Once trained, the predicted kNDVIs were evaluated over the training, validation, and test sets using several performance metrics on a per-cube basis. Only the last year of data (unseen by the model) was used for the validation and test sets, with the test set remaining completely unseen until the end to serve as a proxy for the actual model's performance. Both cloud-covered/unavailable pixels and pixels belonging to a non-vegetation land cover were ignored, as we focused on vegetation impacts. The final value for the metric was computed as the grand mean of the per-cube metrics. The considered metrics were the following: $L_1$ (mean absolute error), $L_2$ (mean square error), $R^2$ score (fraction of kNDVI variance explained by the model), Normalized Nash-Sutcliffe Efficiency (NNSE, used to assess the predictive skill of hydrological models, and equivalent to $\frac{1}{2-R^2}$), and bias (average systematic absolute error committed by the model).












Additionally, the results from all the previous metrics were compared against the results of two \emph{naïve} (parameter-less) models: the climatology at the prediction timestep and the last non-cloud-covered available value. As shown in Table~\ref{tab:results}, these simple models constitute a strong baseline. 

\subsection{Explainable AI for extreme event understanding}\label{sec:explainable-ai}

For obtaining both local and global explanations, we used Python's Captum library \cite{Kokhlikyan2020}, a well-known open-source library for xAI built on PyTorch implementing SOTA attribution methods. After some initial testing on a variety of methods, including Input $\times$ Gradient \cite{Shrikumar2016} and Gradient SHAP \cite{Lundberg2017}, finally Integrated Gradients (IG) \cite{Sundararajan2017} (computed over nine integration steps) was used for the experiments here presented. In IG, attributions are calculated as the integral of the gradients of the outputs with respect to the inputs along the path from a given baseline (e.g., 0) to the input, hence solving (theoretically at least) the issues of sensitivity and implementation invariance of other methods. Out of all the output classes (\emph{B02\_next}, \emph{B03\_next}, \emph{B04\_next}, and \emph{B8A\_next}), attributions were only calculated with respect to \emph{kNDVI\_next}, since the focus of the xAI stage is to analyze the impacts on vegetation. 

\subsubsection{Local and global attributions}\label{sec:local-global}
A naïve use of attribution methods for our problem would yield intractably large attribution tensors since an attribution score is produced for every input and output value combination. Therefore, some local aggregations were proposed. More precisely, the output's spatio-temporal dimensions were always aggregated by taking the mean over some selected timesteps. While some local attribution tensors were represented as is, global attributions were additionally obtained by taking the mean of the input's spatio-temporal dimensions over several minicubes. This procedure is further clarified hereafter in Section \ref{sec:heatwave}.


\subsubsection{The October 2020 central South America heatwave}\label{sec:heatwave}
The xAI analysis in this paper was focused on the October 2020 Central South America heatwave, which affected a large geographical extension (from southern Peruvian Amazon to southeastern Brazil) for a long time (September 23rd to October 15th) and had a strong impact in the region (reaching record temperatures of $10^\circ$C above normal, some locations reporting maximum temperatures above $40^\circ$C for several days in a row). A persistent atmospheric blocking caused the heatwave: a warm air mass lingered for several days, leading to significant temperature anomalies, likely exacerbated by very low soil moisture, as solar energy heated the atmosphere rather than evaporating the non-existent water, which, in turn, worsened drought conditions, escalating fires and impacting natural and human systems \cite{Marengo2022}.

This compound drought and heatwave event had been selected for building the \dataset~dataset (and labeled with event IDs 32379 and 31833), making it ideal for our study. For the xAI analysis of this event, minicubes undergoing event ID 31833, for which the event lasted at least ten days (two timesteps), were selected (and relegated to the test set, see Figure~\ref{fig:map}), for a total of 24 minicubes. Then, the average attributions of these minicubes were obtained for the model's outputs over the event and for the same period but exactly one year before, allowing us to compare feature importances for event and non-event predictions. Global attributions were obtained by taking the mean over the input's spatio-temporal dimensions and over all 24 minicubes (during the event, and one year before the event, see Figure \ref{fig:xai_single_t}). Finally, one of the minicubes was selected, and the complete disaggregated local attributions were analyzed (see Figure \ref{fig:xai_single_txy}).


\section{Results and discussion}\label{sec:results}

\newcommand{\sbullet}{~{\tiny$^\bullet$}~}

\begin{table*}[!tb]
    \centering
    \renewcommand{\arraystretch}{1.2}
    \caption{Results for kNDVI prediction (only over vegetation land covers) for the trained \emph{convLSTM}, the naïve models (\emph{climatology} and \emph{last-value}), and for the ablated models (\emph{LSTM} and \emph{Conv}), for a variety of metrics, over the training, validation, and test sets. In \emph{extremes} / \emph{non-extremes} rows, only timesteps not labeled / labeled as extreme events were used for the evaluation. Relative improvements with respect to the climatology are given in parenthesis. Best general model performances are in bold}

    \begin{tabular}{llccccc}
        {Split (samples)} & 
        {Model (parameters)} & 
        \makecell{\textbf{$L_1$ (improv. \%)}\\ (lower is better)} & 
        \makecell{\textbf{$L_2$ (improv. \%)}\\ (lower is better)} & 
        \makecell{\textbf{$R^2$ (improv. \%)}\\ (higher is better)} & 
        \makecell{\textbf{NNSE (improv. \%)}\\ (higher is better)} & 
        \makecell{\textbf{Bias (improv. \%)}\\ (lower is better)} \\
        
        \toprule
        
        \multirow{5}{*}{\makecell{Training\\($N=27379$)}} 
        & {Climatology (0)} 
        & 0.0739 (0\%) & 0.1214 (0\%) & 0.7251 (0\%) & 0.6351 (0\%) & 0.0225 (0\%) \\
        & {Last-value (0)} 
        & 0.0575 (-22.19\%) & 0.1100 (-9.39\%) & 0.7749 (6.87\%) & 0.6699 (5.48\%) & \textbf{0.0028 (-87.56\%)} \\
        \cline{2-7}

        & {ConvLSTM (768k)} 
        & \textbf{0.0422 (-42.90\%)} & \textbf{0.0792 (-34.76\%)} & \textbf{0.8820 (21.74\%)} & \textbf{0.7888 (24.20\%)} & 0.0088 (-60.89\%) \\
        \cline{2-7}

        & {LSTM (495k)} 
        & 0.0442 (-40.19\%) & 0.0844 (-30.48\%) & 0.8671 (19.58\%) & 0.7780 (22.62\%) & 0.0172 (-23.55\%) \\
        & {Conv (744k)} 
        & 0.0529 (-28.41\%) & 0.0934 (-23.06\%) & 0.8373 (15.47\%) & 0.7395 (16.44\%) & 0.0123 (-45.33\%) \\
        \midrule

        \multirow{9}{*}{\makecell{Validation\\($N=3644$)}}
        & {Climatology (0)} 
        & 0.0688 (0\%) & 0.1103 (0\%) & 0.7731 (0\%) & 0.6581 (0\%) & 0.0273 (0\%) \\
        & {\sbullet Extremes}
        & 0.0767 (11.46\%) & 0.1067 (-3.26\%) & 0.7288 (-5.73\%) & 0.5212 (TODO\%) & 0.0471 (TODO\%) \\
        
        & {Last-value (0)} 
        & 0.0561 (-18.46\%) & 0.1079 (-2.18\%) & 0.7827 (1.24\%) & 0.6843 (3.98\%) & \textbf{0.0025 (-90.84\%)} \\
        & {\sbullet Extremes}
        & 0.0589 (-14.39\%) & 0.0953 (-13.60\%) &  0.7838 (1.38\%) & 0.6182 (-20.80\%) & 0.0364 (33.33\%) \\
        \cline{2-7}
        
        & {ConvLSTM (768k)} 
        & \textbf{0.0413 (-39.97\%)} & \textbf{0.0791 (-28.29\%)} & \textbf{0.8831 (14.23\%)} & \textbf{0.7852 (19.32\%)} & 0.0114 (-58.24\%) \\
        & {\sbullet Non-extremes} 
        & 0.0413 (-39.97\%) & 0.0791 (-28.29\%) & 0.8831 (14.23\%) & 0.7851 (19.30\%) & 0.0114 (-58.24\%) \\
        & {\sbullet Extremes}
        & 0.0482 (-29.94\%) & 0.0781 (-29.19\%) & 0.8546 (10.54\%) & 0.6716 (2.05\%) & 0.0340 (24.54\%) \\
        \cline{2-7}
        
        & {LSTM (495k)} 
        & 0.0420 (-38.95\%) & 0.0820 (-25.66\%) & 0.8747 (13.14\%) & 0.7748 (17.73\%) & 0.0181 (-33.70\%) \\
        & {Conv (744k)} 
        & 0.0505 (-26.70\%) & 0.0897 (-18.68\%) & 0.8499 (9.93\%) & 0.7379 (12.23\%) & 0.0146 (-46.52\%) \\
        \midrule

        \multirow{7}{*}{\makecell{Test\\($N=3678$)}} 
        & {Climatology (0)} & 0.0658 (0\%) & 0.1065 (0\%) & 0.7830 (0\%) & 0.6843 (0\%) & 0.0249 (0\%) \\
        & {Last-value (0)} 
        & 0.0480 (-27.05\%) & 0.0939 (-11.83\%) & 0.8311  (6.14\%) & 0.7478 (9.28\%) & \textbf{0.0016 (-93.57\%)} \\
        \cline{2-7}
        
        & {ConvLSTM (768k)} 
        & \textbf{0.0364 (-44.68\%)} & \textbf{0.0702 (-34.08\%)} & \textbf{0.9055 (15.64\%)} & \textbf{0.8321 (21.60\%)} & 0.0076 (-69.48\%) \\
        & {\sbullet Non-extremes} 
        & 0.0363 (-44.83\%) & 0.0702 (-34.08\%) & 0.9054 (15.63\%) & 0.8321 (21.60\%) & 0.0076 (-69.48\%) \\
        & {\sbullet Extremes}
        & 0.0270 (-58.97\%) & 0.0411 (-61.41\%) & 0.9614 (22.78\%) & 0.8327 (21.69\%) & 0.0199 (-20.08\%) \\
        \cline{2-7}

        & {LSTM (495k)} 
        & 0.0370 (-43.77\%) & 0.0721 (-32.30\%) & 0.9003 (14.98\%) & 0.8231 (20.28\%) & 0.0125 (-49.80\%) \\        
        & {Conv (744k)} 
        & 0.0458 (-30.39\%) & 0.0823 (-22.72\%) & 0.8703 (11.15\%) & 0.7807 (14.09\%) & 0.0131 (-47.39\%) \\
        \bottomrule
        \label{tab:results}
    \end{tabular}
\end{table*}

\subsection{Performance evaluation}\label{sec:forecasting-model}

\begin{figure*}[!tb]
    \centering
    \includegraphics[width=\linewidth]{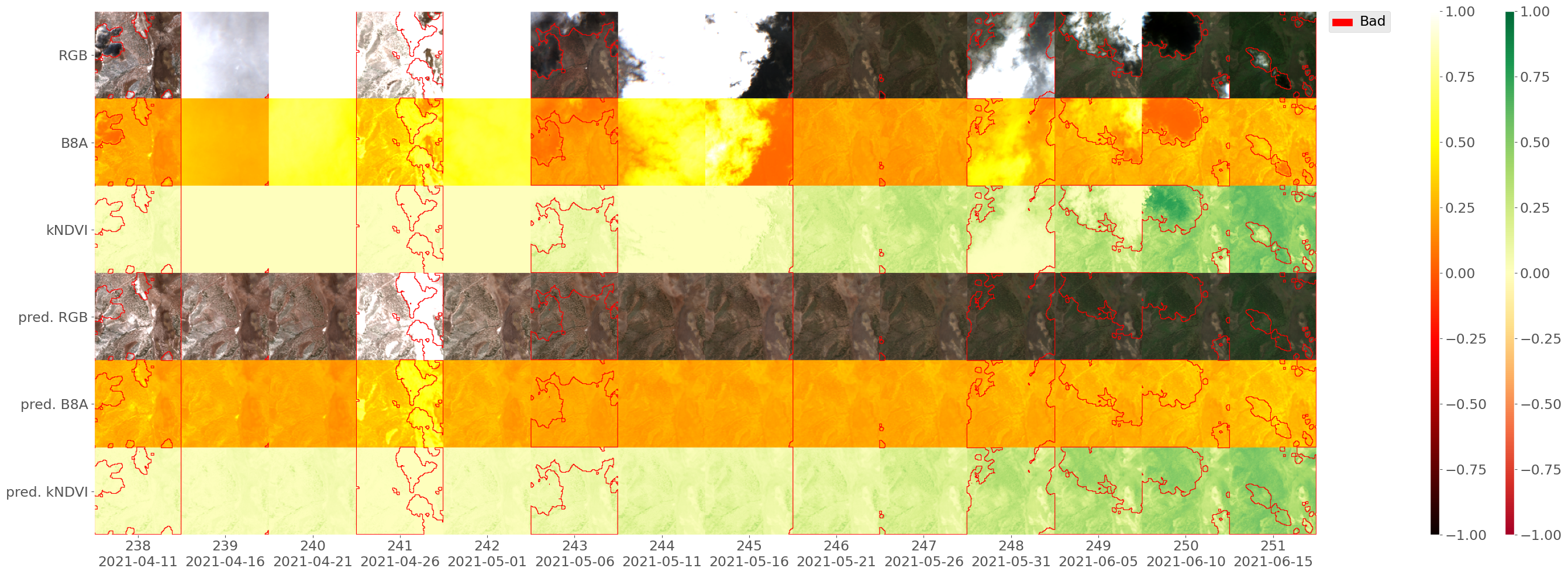}
    \caption{A Minicube at location 101.95\textdegree{}E, 46.97\textdegree{}N (Mongolia), from 2021-04-11 to 2021-06-15. Top three rows: ground truth \emph{RGB\_next}, \emph{B8A\_next}, and \emph{kNDVI\_next}. Bottom three rows: model predictions. Red outline: \emph{cloud\_mask} (labeled as "bad"). Minicube $L_1$: 0.0351, $L_2$: 0.0587, $R^2$: 0.9288, NNSE: 0.9335, bias: 0.0095}.
    \label{fig:result}
\end{figure*}

The quantitative results of the forecasting model are summarized in Table~\ref{tab:results}. As can be seen, the convLSTM outperforms the baseline models (climatology and last-value) to a great extent (20-40\% in all metrics except for the bias) over the training, validation, and test sets, meaning that the model has successfully learned to generalize. Although absolute scores seem better in the test set, relative improvements are similar to other splits, possibly suggesting that this set must contain cubes that are easier to predict. 

Comparing against the ablated models (\textit{LSTM} and \textit{Conv}), the convLSTM seems to perform better, suggesting that the model benefits both from the spatial and the temporal context. Also, the performance of the convLSTM falls significantly when evaluated only on timesteps labeled as extreme events, which is expected, as they should be anomalous and, hence, more challenging to predict. This performance drop is also observed in the climatology baseline (since an extreme is a deviation from climatology) but not as much in the last-value baseline. Curiously, the convLSTM performance in the test set is excellent in the presence of extremes, which justifies using xAI over these events in Section~\ref{sec:results-explainable-ai}. It can also be seen that the performance in non-extreme timesteps is the same as the global performance due to the tiny proportion of extreme timesteps.

For a qualitative assessment of the results, Figure \ref{fig:result} shows the model's predictions along with the Ground Truth (GT) reflectance values for a few timesteps of a single test minicube. On the one hand, for the first few timesteps, the model accurately predicts the progressive thawing of the snow and the sudden snowfall present in the fourth timestep (despite the two previous timesteps being entirely covered by snow). On the other hand, the convLSTM correctly predicts the gradual greening of the scene over the latter timesteps. Furthermore, the forecasts seem robust to cloud coverage while doing a reasonable job at gap filling even if not explicitly trained for the task.

\subsection{Explainable AI for the October 2020 central South America heatwave}\label{sec:results-explainable-ai}

\begin{figure}[!ht]
    \centering
    \includegraphics[width=\columnwidth]{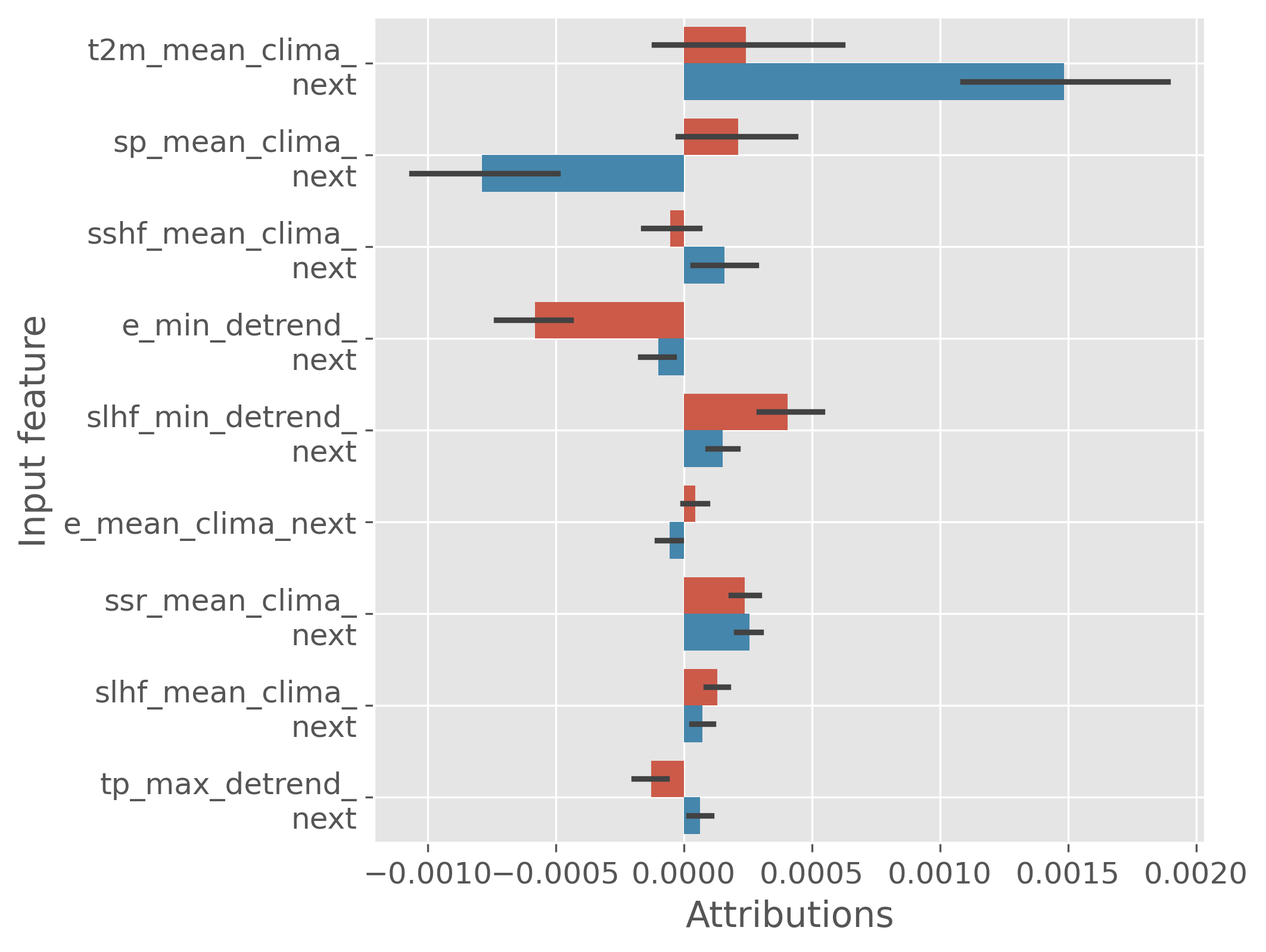}
    \includegraphics[width=\columnwidth]{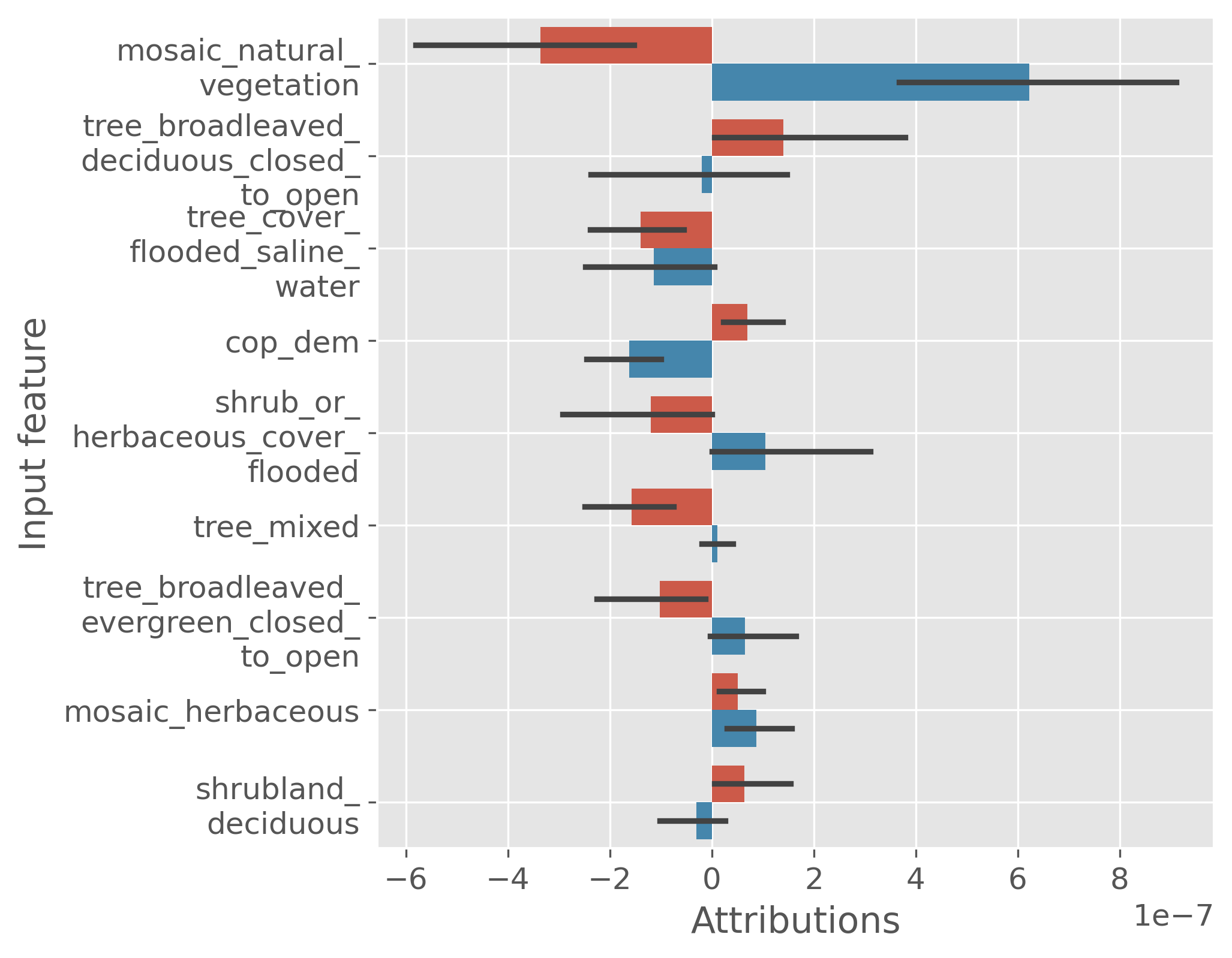}
    \includegraphics[width=\columnwidth]{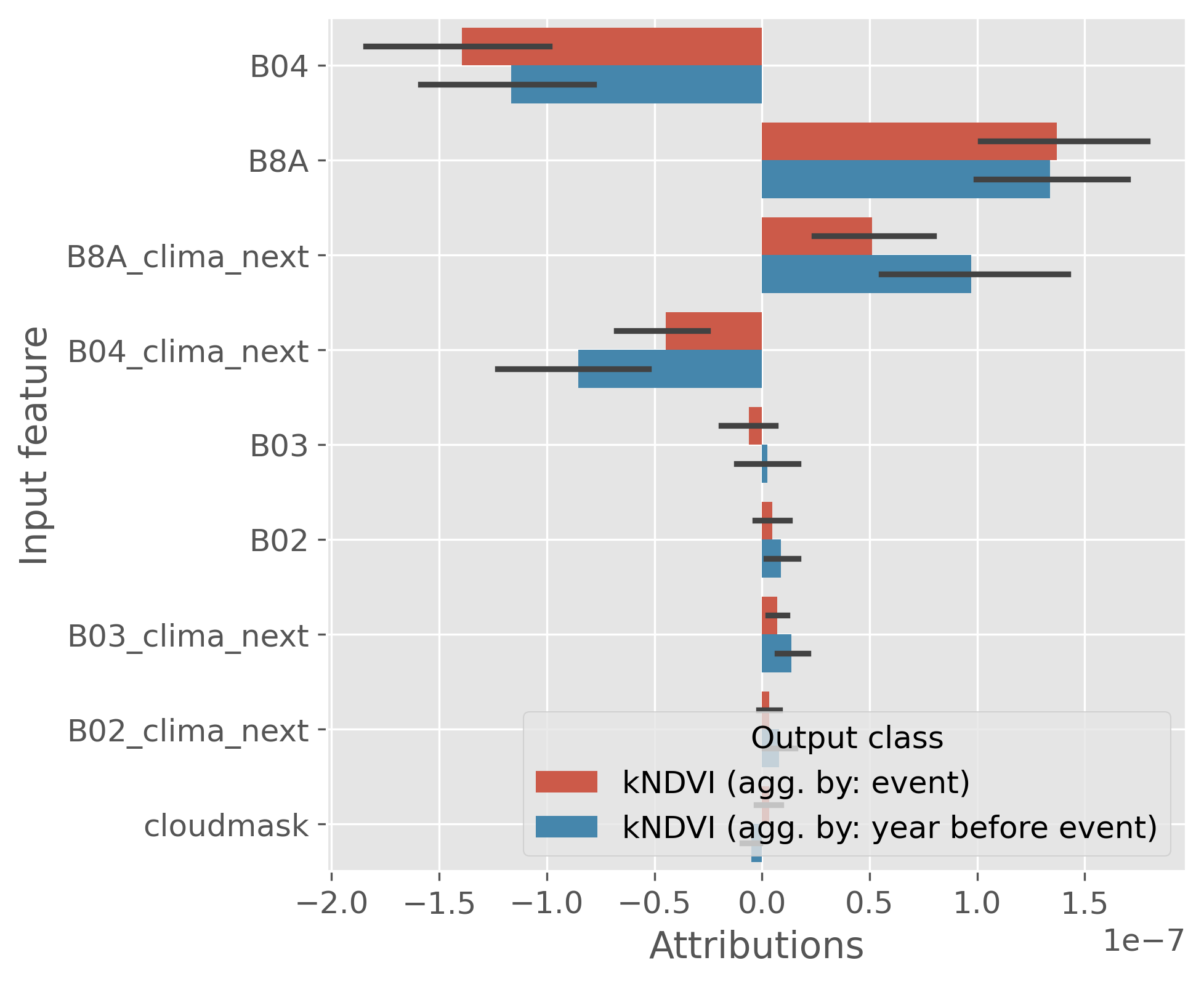}
    \caption{Bar plot of the average attributions (over time, space, and minicubes) for top nine variables in tensor $x_{t}$ (top), $x_{s}$ (middle), and $x_{st}$ (bottom) of 24 minicubes affected by event 31833 (October 2020 central South America heatwave). Red bars represent the attribution for the model's outputs coinciding with the event, while the blue bars represent the attributions for the same period but one year before.}
    \label{fig:xai_event}
\end{figure}

In Section~\ref{sec:forecasting-model}, the convLSTM was shown to have strong performance for vegetation impact forecasting (kNDVI) under extreme events in the test set, which is a prerequisite for any later xAI analysis to be useful (i.e., there are no insights to be gained from a model that performs poorly). Figure~\ref{fig:xai_event} shows the average attributions (over time, space, and minicubes) for variables in tensors $x_t, x_s$ and $x_{st}$ of minicubes affected by the October 2020 central South America heatwave, with red bars representing the attribution for the event, and the blue bars representing the attributions for the same period, but one year before. 

Analyzing $x_{t}$ (ERA5-derived) variables, when there is no event happening, the two most important ones are average climatology (\emph{\_mean\_clima}) of the temperature at 2 meters (\emph{t2m}) and the surface pressure (\emph{sp}) and, within events, the minimum anomaly (\emph{\_min\_detrend}) of the evaporation (\emph{e}) and of the surface latent heat flux (\emph{slhf}). As is perhaps expected, the model pays more attention to instantaneous values (or anomalies in this case) rather than climatology to predict vegetation impacts under extremes. Also, \emph{t2m} and \emph{sp} seem to be the best general predictors for kNDVI (outside extremes), with higher average temperatures leading to higher kNDVI values and, conversely, higher pressures resulting in lower kNDVI values. During extremes, however, minimum \emph{e} (negative sign indicates increased evaporation) and \emph{slhf} (positive sign indicates vertical flux downwards) are the most relevant predictors. Yet, the sign of their contributions is much more challenging due to the minimum operator.

Regarding $x_{s}$ variables (land cover and DEM), the most important landcover class seems to be \textit{mosaic\_natural\_vegetation}, which has a considerable positive attribution on kNDVI outside of extremes, but a negative one during them, perhaps meaning that, once the natural vegetation is dead as a result of the heatwave, it becomes a negative predictor for kNDVI. Not much else can be said for the other variables due to the large confidence bars. Finally, for $x_{st}$ (reflectance-derived and cloud mask) variables, \emph{B04} and \emph{B8A} have the highest attribution (which is expected since kNDVI is directly computed from them), followed by their corresponding climatology values (\emph{B04\_clima\_next} and \emph{B8A\_clima\_next}). The sign of the attributions even corresponds with the sign that these bands (\emph{B04} and \emph{B8A}) have in the numerator of the kNDVI equation (\ref{eq:kndvi}). Similar to what was observed for $x_{t}$, the importance of climatology is much lower during extremes, where conditions are, by definition, exceptional. Interestingly, despite the model's ability to gap-fill cloud-covered regions, the cloud mask itself has very little use for the model.

\begin{figure*}[tb]
    \centering
    \includegraphics[width=\textwidth]{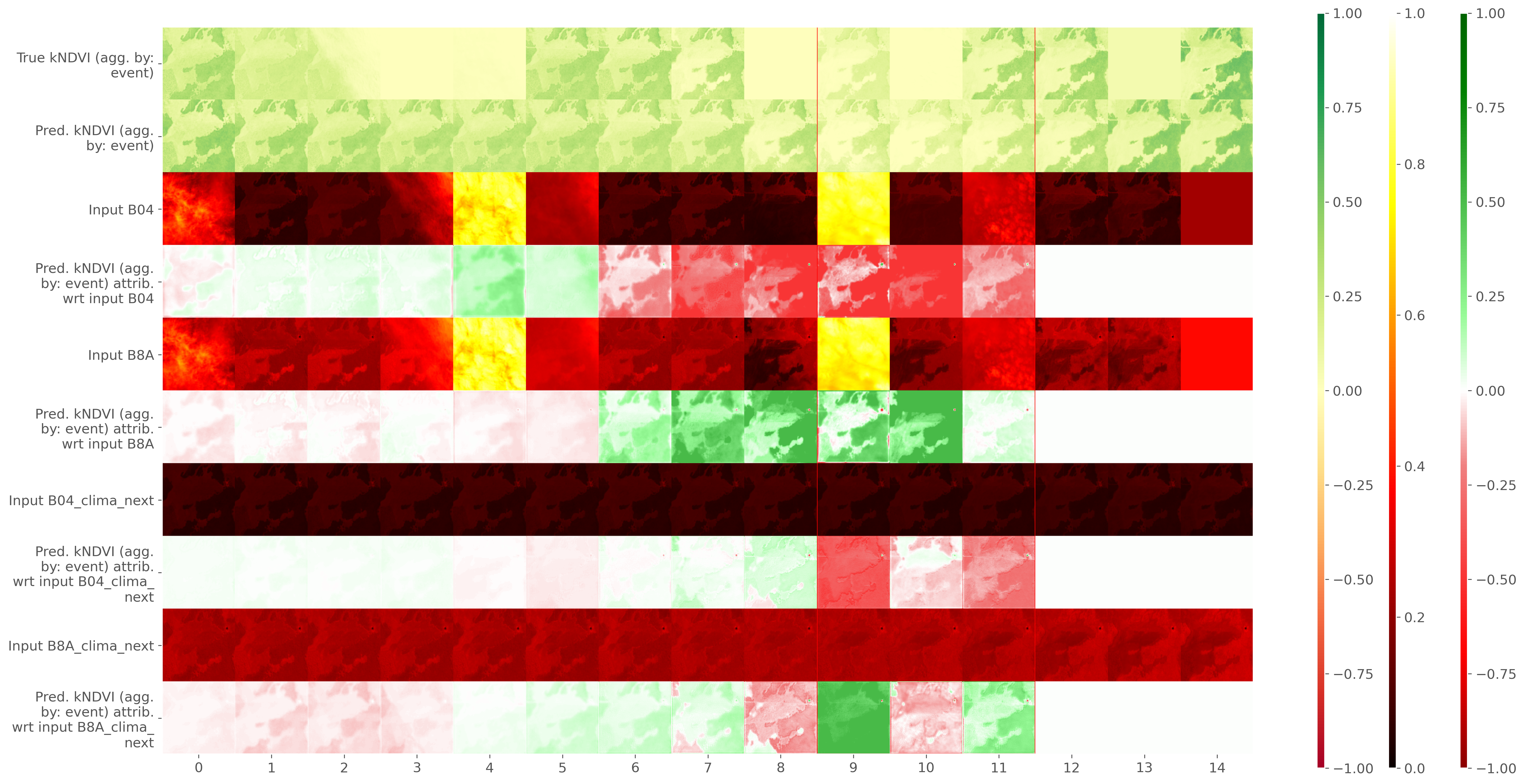}
    \caption{Full attributions for a few timesteps of input $x_{st}$ from a minicube sampled at -58.17\textdegree{}E -23.98\textdegree{}N (Paraguay) and affected by event 31833 (October 2020 central South America heatwave). The first two rows represent the ground truth kernel Normalized Vegetation Index (kNDVI) and the model's prediction, respectively. In contrast, the rest of the rows represent an input feature (only the top four by importance) and its corresponding attribution map (green: positive, red: negative) in an alternating pattern. Thus, the first two rows are offset by one timestep into the future with respect to the rest. There is a red outline around timesteps 9-11, signaling where event 31833 occurred (2020-10-03, 2020-10-08, and 2020-10-13). Attributions were rescaled to lie within the $[-1,1]$ range, while remaining zero-centered.}
    \label{fig:xai_single_txy}
\end{figure*}

\begin{figure}[!ht]
    \centering
    \includegraphics[width=\columnwidth]{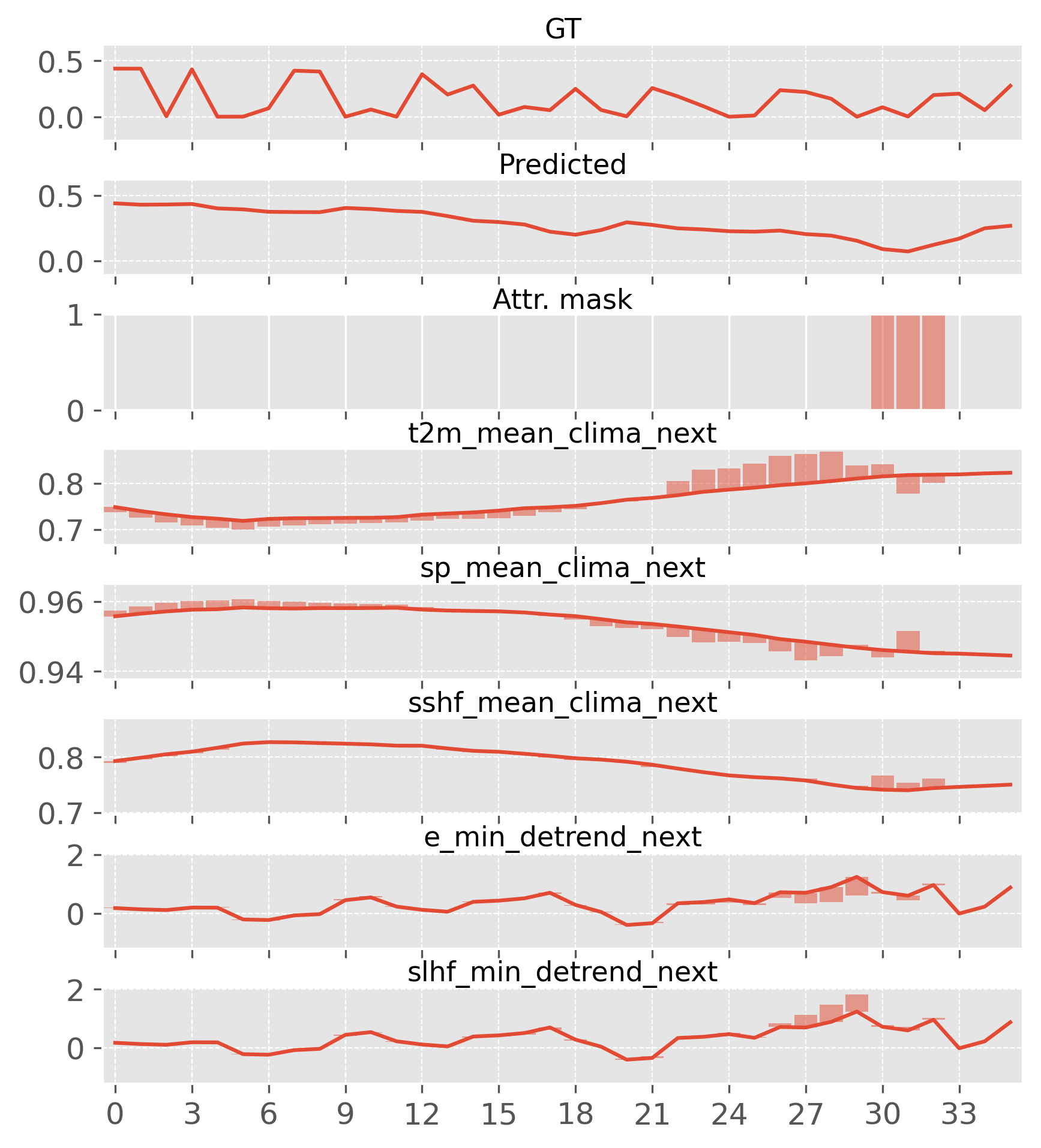}
    \caption{Full attributions for a few timesteps of input $x_{t}$ from a minicube sampled at -58.17 -23.98 (Paraguay) and affected by event 31833 (October 2020 Central South America heatwave). The first three plots represent, respectively, the average ground truth (GT) kNDVI, the predicted kNDVI, and a mask indicating the timesteps (30-32, equivalent to 2020-10-03, 2020-10-08, and 2020-10-13) over which the attribution (and also the event) took place. In contrast, the rest of the plots show the top five most important input feature values (as a line) and their corresponding attributions (as a bar: positive when going upwards and negative otherwise). }
    \label{fig:xai_single_t}
\end{figure}

Figures~\ref{fig:xai_single_txy} and \ref{fig:xai_single_t} show the full attribution tensors for inputs $x_{st}$ and $x_{t}$, respectively, from a minicube affected by event 31833 (October 2020 central South America heatwave). The first two rows of Figure~\ref{fig:xai_single_txy} represent the GT kNDVI and the model's prediction, whereas the rest of the rows represent an input feature and its corresponding attribution map in an alternating pattern. Thus, the first two rows are offset by one timestep into the future with respect to the rest. There is a red outline around timesteps 9-11, signaling where event 31833 occurred (2020-10-03, 2020-10-08, and 2020-10-13). Firstly, we see that the model accurately predicted the reduction in greenness during the extreme event. Similar to the conclusions of the previous section, \emph{B04} and \emph{B8A} (as well as their climatology) seem to have opposite attributions with a sign that is generally explained by the kNDVI equation (\ref{eq:kndvi}). However, this figure shows a much greater complexity: the sign of the attributions is inverted at around timesteps 5-6, perhaps representing a change of regime as the extreme started to develop (which might be helpful to assess for how long the event was brewing); the attributions become smaller as we move back in time from the attributed timestep (although they never quite reach zero if we could look at the full image), and they are precisely zero after the attribution timesteps (since the convLSTM model is sequential and cannot look into the future); cloud-occluded reflectances have a strong attribution (opposite to the idea that they should not contribute), perhaps meaning that the presence of clouds is itself predictive of future kNDVIs; different parts of a single tile contribute differently to the attribution; etc. Unfortunately, the complexity makes the analysis extremely hard, supporting the use of aggregations of the previous study.

Similarly, the first three plots in Figure~\ref{fig:xai_single_t} represent, respectively, the average Ground Truth (GT) kNDVI, the predicted kNDVI, and a mask indicating the timesteps (30-32, equivalent to 2020-10-03, 2020-10-08, and 2020-10-13) over which the attribution (and also the event) took place; whereas the rest of the plots show the input feature values as a line, along with their corresponding attributions as a bar (positive when going upwards, and negative otherwise). Compared to Figure~\ref{fig:xai_single_txy}, a longer context was chosen for this figure. Firstly, we see that the average predicted kNDVI (second plot) goes down progressively over time, reaching a minimum that coincides with the event; this cannot be as easily appreciated in the GT signal due to the perpetual presence of clouds. Furthermore, climatology variables' attributions go much farther back than the instantaneous variables, which are only relevant for the last few timesteps before the prediction, helping explain why the climatology variables tend to have higher attributions when averaged over time. Similar to Figure~\ref{fig:xai_single_txy}, the climatology variables seem to change regime during the extreme, with temperature stopping being positively related to kNDVI and surface pressure starting to do so. Conversely, instantaneous variables are important right before the event but, surprisingly, lose their relevance during the event itself.

\section{Conclusion}

The successful marriage of high-dimensional EO data modeling and xAI can help us predict and understand the occurrence of extreme events affecting ecosystems, mainly compound heatwave and drought events, which are particularly interesting due to their vast impacts. This study provides the tools to make it possible through the accurate forecasting of kNDVI using DL models, the explanation of these models' behavior in the presence of such events, and their visualization.

From the data processing point of view, an efficient and causal climatology was employed as direct input to the model and for detrending other variables, keeping only the anomaly signal. Both variables were later proven helpful in the xAI analysis, climatology being important during non-extremes and anomalies during extremes. From the modeling side, predicting only the anomalies in reflectance with respect to the climatology helped the model converge, while introducing the kNDVI computation in the loss (instead of predicting it directly) helped the model efficiently learn both reflectances and kNDVI. From the explainability point of view, Captum's attribution methods have been adapted to spatio-temporal data, along with the tools to visualize the explanations, allowing us to gain interesting insights that substantiate the model's behavior during extremes.

However, some significant limitations remain: the model predicts only one timestep ahead, the provided explanations could not always be linked satisfactorily to the domain knowledge, and the data handling is computationally costly. For future research, some promising avenues include: the use of xAI storylines for extreme event understanding, the exploration of other xAI aggregation methods beyond simple averaging (e.g. PCA, clustering), or the training of more modern (e.g. attention-based) architectures. By making the code publicly available, together with the public \dataset~dataset, we hope to encourage the community to use these tools and baseline models to further improve Earth surface forecasting ability and extreme event understanding, widening this, so far, widely unexplored field.

\section*{Acknowledgment}
This work was supported by the ESA AI4Science project "Multi-Hazards, Compounds and Cascade events: DeepExtremes," 2022-2024; the European Union’s Horizon 2020 research and innovation project XAIDA: Extreme Events - Artificial Intelligence for Detection and Attribution, 2021-2024 (grant agreement No 101003469); and the European Union's Horizon 2022 project ELIAS: European Lighthouse of AI for Sustainability, 2023-2027 (grant agreement No. 101120237).

\bibliographystyle{IEEEtran}
\bibliography{main}


\end{document}